\documentclass[letterpaper]{article} 
\usepackage[submission]{aaai25}  
\usepackage{times}  
\usepackage{helvet}  
\usepackage{courier}  
\usepackage[hyphens]{url}  
\usepackage{graphicx} 
\usepackage{natbib}  
\usepackage{caption} 
\usepackage{algorithm}

\usepackage[utf8]{inputenc} 
\usepackage[T1]{fontenc}    

\usepackage{url}            
\usepackage{booktabs}       
\usepackage{amsfonts}       
\usepackage{nicefrac}       
\usepackage{microtype}      
\usepackage{xcolor}         

\usepackage{graphicx}
\usepackage{amsmath}
\usepackage{makecell}
\usepackage{multirow}
\usepackage{algorithm}
\usepackage{amssymb}
\usepackage{verbatim}
\usepackage{graphicx}
\usepackage{subfig}

\usepackage{algpseudocode}
\usepackage{color}
\usepackage{algorithm}
\usepackage{float}

\usepackage{newfloat}
\usepackage{listings}
\DeclareCaptionStyle{ruled}{labelfont=normalfont,labelsep=colon,strut=off} 
\floatstyle{ruled}
\newfloat{listing}{tb}{lst}{}
\floatname{listing}{Listing}
\title{Scene123: One Prompt to 3D Scene Generation  via Video-Assisted and Consistency-Enhanced MAE}
\author {
    Yiying Yang\textsuperscript{\rm 1}$^*$, Fukun Yin\textsuperscript{\rm 2}\thanks{Yiying Yang and Fukun Yin contributed equally to this work.}, 
    Jiayuan Fan\textsuperscript{\rm 1}\thanks{ Corresponding author.},
    Wanzhang Li\textsuperscript{\rm 2},
    Xin Chen\textsuperscript{\rm 3},
    Gang Yu\textsuperscript{\rm 3}
}
\affiliations {

\normalsize
    $^1$ Academy for Engineering and Technology, Fudan University\\
    $^2$ School of Information Science and Technology, Fudan University \\
    $^3$ Tencent PCG\\
 {  yiyingyang23@m.fudan.edu.cn, fkyin21@m.fudan.edu.cn, jyfan@fudan.edu.cn, \\  liwz22@m.fudan.edu.cn, chenxin2@shanghaitech.edu.cn, iskicy@gmail.com } 
}

\usepackage{bibentry}

\begin{document}

\maketitle

\begin{abstract}
 As Artificial Intelligence Generated Content (AIGC) advances, a variety of methods have been developed to generate text, images, videos, and 3D 
 shapes from single or multimodal inputs, contributing efforts to emulate human-like cognitive content creation. 
 However, generating realistic large-scale scenes from a single input presents a challenge due to the complexities involved in ensuring consistency across extrapolated views generated by models.
 Benefiting from recent video generation models and implicit neural representations, we propose Scene123, a 3D scene generation model, which combines a video generation framework to ensure realism and diversity with implicit neural fields integrated with Masked Autoencoders (MAE) to effectively ensure the consistency of unseen areas across views.
Specifically, the input image (or a text-generated image) is first warped to simulate adjacent views, with the invisible regions filled using the consistency-enhanced MAE model. 
 Nonetheless, the synthesized images often exhibit inconsistencies in viewpoint alignment, thus we utilize the produced views to optimize a neural radiance field, enhancing geometric consistency.
 Moreover, to further enhance the details and texture fidelity of generated views, we employ a GAN-based Loss against images derived from the input image through the video generation model.
%
 Extensive experiments demonstrate that our method can generate realistic and consistent scenes from a single prompt. Both qualitative and quantitative results indicate that our approach surpasses existing state-of-the-art methods. We show encourage video examples at  \url {https://yiyingyang12.github.io/Scene123.github.io/ }.
 
\end{abstract}

\begin{figure}[htbp]
  \centering
  \includegraphics[width=0.98\linewidth]{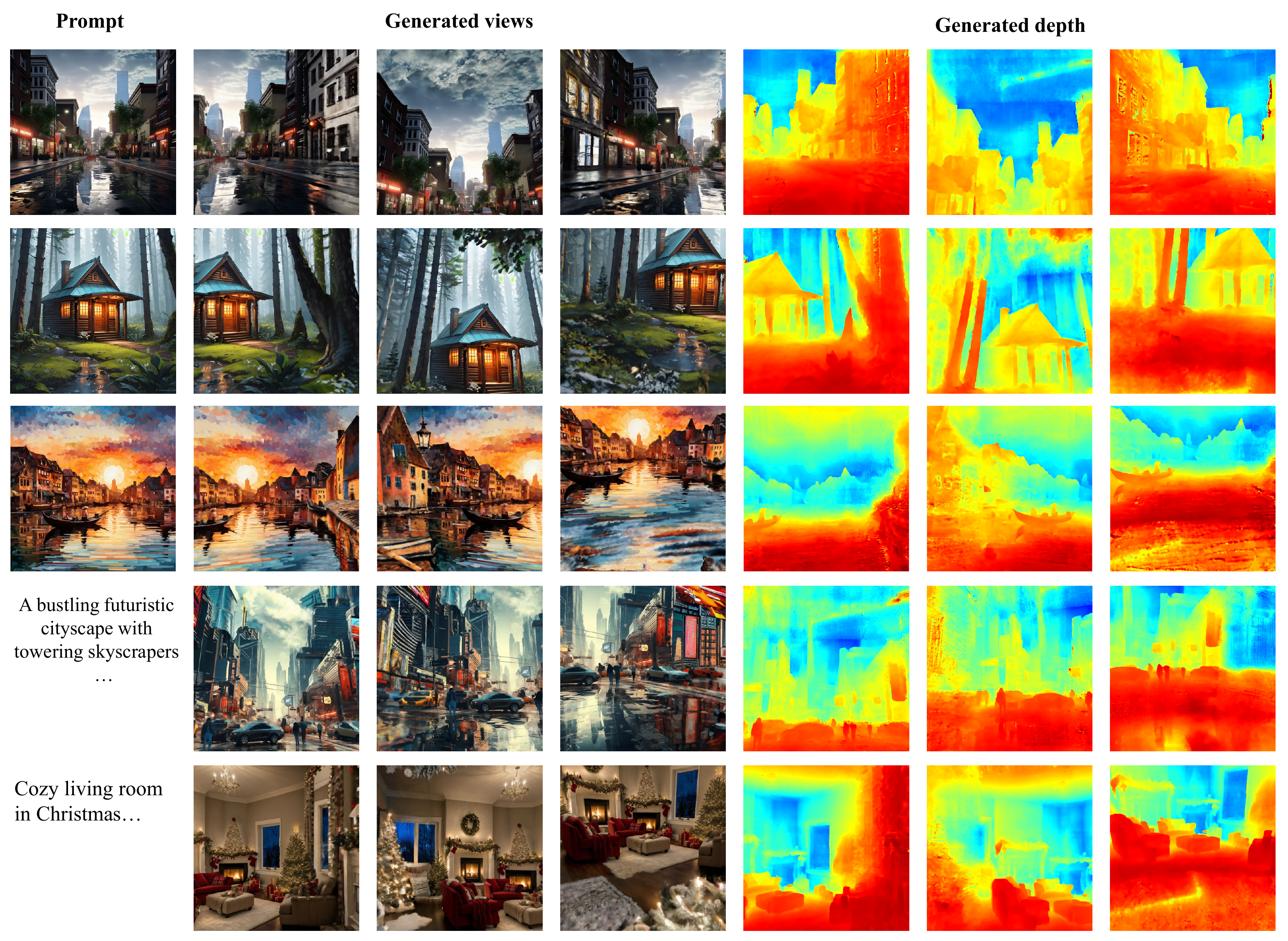}
  \vspace{-2mm}
  \caption{Some examples generated by our Scene123. For a single input image or text, our method can generate 3D scenes with consistent views, fine geometry, and realistic textures, applicable to real, virtual, or object-centered scenes.}
  \vspace{-3mm}
  \label{fig:teaser}
\end{figure}

\section{Introduction}
3D scene generation aims to create realistic or stylistically specific scenes from limited prompts, such as a few images or a text description. This is a fundamental issue in computer vision and graphics and a critical challenge in generative artificial intelligence.
Recent advancements have demonstrated substantial progress through the use of vision-language models~\cite{radford2021learning}, generative models such as Generative Adversarial Networks (GANs)~\cite{zhang2019stylistic}, Variational autoencoders (VAEs)~\cite{sargent2023vq3d,yang2023vq}, or diffusion models~\cite{yang2023law}, and scene representations like Neural Radiance Fields (NeRF)~\cite{mildenhall2021nerf,yin2022coordinates,ding2023pdf,lu2023large,yang2024pm} or 3D Gaussian Splatting (3DGS)~\cite{kerbl20233d}. 
Typically, these methods begin by generating images with a pretrained generative model~\cite{rombach2022high}, or by directly using images for image-to-scene generation, then estimate additional 3D surface geometric details such as depth and normal~\cite{piccinelli2023idisc, li2023neaf, yin2023dcnet, yin2020accurate}, and subsequently render the surface textures of the scenes using generative strategies like inpainting~\cite{wang2023imagen}.
Moreover, some approaches update the geometric surfaces as new views are generated to maintain the coherence of the scene~\cite{zhang2024text2nerf}.
However, these methods often rely on pre-trained models~\cite{radford2021learning, li2022blip}, resulting in inconsistencies and artifacts in the generated scenes. Additionally, these methods face challenges in producing high-quality, coherent 3D representations across diverse and complex environments.

In this paper, we endeavor to address this challenge about generating 3D scenes from a single image or textual description, ensuring viewpoint consistency and realistic surface textures for both real and synthetically styled scenes, as shown in Fig. \ref{fig:teaser}.
However, achieving a balance between viewpoint consistency and flexibility presents a significant challenge. View consistency 
requires maintaining coherent and accurate details across multiple perspectives, which can limit the model's adaptability to diverse inputs and tasks.
Conversely, flexibility requires the model to produce high-quality outputs under varying conditions, which may introduce inconsistencies in the generated scenes.

It is noteworthy that scene synthesis based on multi-view images is a long-studied topic from the early Multi-View Stereo (MVS) ~\cite{schoenberger2016mvs, schoenberger2016sfm} to recent implicit neural representations~\cite{yu2021pixelnerf}, ensuring view consistency in generated scenes through multi-view matching mechanisms. 
However, when the input is reduced to a single image or a textual description, lacking reference multi-view images, the performance of these methods is greatly compromised.
Fortunately, methodologies such as Masked Autoencoders (MAE) ~\cite{he2022masked} provide avenues for extrapolating to areas unseen in new views, while the incorporation of additional semantic layers augments the coherence of the synthesized scenes. Additionally, video generation models ~\cite{luo2023videofusion} facilitate the enhancement of scenes with richer priors and more detailed texture information. Thus, how to utilize multi-view reconstruction with robust physical constraints alongside effective expansions in new viewpoints and video generation models is still an unreached area.

Scene123 investigates a methodology for 3D scene reconstruction employing stringent physical constraints alongside a robust multi-view MAE and video generation models to ensure view consistency. 
To achieve this, for each input or generated single image, we first create images of nearby perspectives through warping. 
Subsequently, a consistency-enhanced MAE is designed to inpaint the unseen areas, with a shared codebook maintained to distribute global information to every invisible area. 
We implement a progressive strategy derived from Text2NeRF ~\cite{zhang2024text2nerf} to incrementally update perspectives throughout this process.
Furthermore, to enhance the scene's detail and realism, we employ the latest video generation technology, generating high-quality scene videos based on input images and performing adversarial enhancements with the rendered images. 
The synergistic operation of these two modules enables Scene123 to generate consistent, finely detailed three-dimensional scenes with photo-realistic textures from a single prompt.

We conduct extensive experiments on text-to-scene and image-to-scene generation, encompassing both real and virtual scenes. The results offer robust empirical evidence that strongly supports the effectiveness of our framework. 
Our contributions can be summarized as:
1) We propose a novel scene generation framework based on one prompt, which establishes the connection between MAE and video generation models for the first time to ensure view consistency and realism of the generated scenes.
2) The Consistency-Enhanced MAE is designed to fill unseen areas in new views by injecting global semantic information and combining it with neural implicit fields, ensuring consistent surface representation across various views.
3) We introduce the video-assisted 3D-aware generative refinement module, which enhances scene reconstruction by integrating the diversity and realism of video generation models through a GAN-based function to significantly improve detail and texture fidelity.
4) Extensive experiments validate the efficacy of Scene123, demonstrating greater accuracy in surface reconstruction, higher realism in reconstructed views, and better texture fidelity compared to the SOTA methods. The data and code will be available.

\section{Related Work}

\textbf{Text to 3D Scene Generation. }
Many recent advancements in text-driven 3D scene generation have focused on modeling 3D scenes using text inputs~\cite{hwang2023text2scene,zhang2024text2nerf}. Due to the scarcity of paired text-3D scene data, most studies utilize Contrastive Language-Image Pre-training (CLIP)~\cite{radford2021learning} or pre-trained text-to-image models to interpret the text input. Text2Scene~\cite{hwang2023text2scene} employs CLIP to model and stylize 3D scenes from text (or image) inputs by decomposing the scene into manageable sub-parts. Set-the-Scene~\cite{cohen2023set} and Text2NeRF~\cite{zhang2024text2nerf} generate NeRFs from text using text-to-image diffusion models to represent 3D scenes. SceneScape~\cite{fridman2024scenescape} and Text2Room~\cite{hollein2023text2room} leverage a pre-trained monocular depth prediction model for enhanced geometric consistency and directly generate the 3D textured mesh representation of the scene. However, these methods depend on inpainted images for scene completion, which, while producing realistic visuals, suffer from limited 3D consistency. More recent studies~\cite{bai2023componerf} have successfully generated multi-object compositional 3D scenes. Another class of methods utilizes auxiliary inputs, such as layouts~\cite{po2023compositional}, to enhance scene generation. Unlike text captions, which can be rather vague, some approaches generate 3D scenes from image inputs, where the 3D scene corresponds closely to the depicted image. Early scene generation methods often require specific scene data for training to obtain category-specific scene generators, such as GAUDI~\cite{bautista2022gaudi}, or focus on single scene reconstruction based on the input image, such as PixelSynth~\cite{rockwell2021pixelsynth} and Worldsheet~\cite{hu2021worldsheet}. However, these methods are often limited by the quality of the generation or the extensibility of the scene. 

\noindent \textbf{Image to 3D Scene Generation. }Recently, numerous studies have concentrated on generating 3D scenes from image inputs. PERF~\cite{wang2023perf} generates 3D scenes from a single panoramic image, using diffusion models to supplement shadow areas. ZeroNVS~\cite{sargent2023zeronvs} extends this capability by reconstructing both objects and environments in 3D from a single image. Despite its environmental reconstruction lacking some detail, the algorithm demonstrates an understanding of environmental contexts. LucidDreamer~\cite{chung2023luciddreamer} and WonderJourney~\cite{yu2023wonderjourney} utilize general-purpose depth estimation models to project hallucinated 2D scene extensions into 3D representations. However, these methods still face challenges in achieving realism, often producing artifacts due to the reliance on pre-trained models.

\noindent \textbf{Video Diffusion Models and 3D-aware GANs.} Diffusion models~\cite{song2020score} have recently emerged as powerful generative models capable of producing a wide array of images~\cite{blattmann2023align} and videos~\cite{blattmann2023stable} by iteratively denoising a noise sample. Among these models, the publicly available Stable Diffusion (SD)~\cite{rombach2022high} and Stable Video Diffusion (SVD)~\cite{blattmann2023stable} exhibit strong generalization capabilities due to training on extremely large datasets such as LAION~\cite{schuhmann2022laion} and LVD~\cite{blattmann2023stable}. Consequently, they are frequently used as foundational models for various generation tasks, including novel view synthesis. To enhance generalization and multi-view consistency, some contemporary works leverage the temporal priors in video diffusion models for object-centric 3D generation. For instance, IM-3D~\cite{melas20243d} and SV3D~\cite{voleti2024sv3d} explore the capabilities of video diffusion models in object-centric multi-view generation. V3D~\cite{chen2024v3d} extends this approach to scene-level novel view synthesis. However, these methods often produce unsatisfactory results for complex objects or scenes, leading to inconsistencies among multiple views or unrealistic geometries.

Early works, like 3D-GAN~\cite{wu2016learning}, Pointflow~\cite{yang2019pointflow}, and ShapeRF~\cite{cai2020learning} focus more on the category-specific texture-less geometric shape generation based on the representations of voxels or point clouds. However, limited by the generation capabilities of GANs, these methods can only generate rough 3D assets of specific categories. Subsequently, HoloGAN~\cite{nguyen2019hologan}, GET3D~\cite{gao2022get3d}, and EG3D~\cite{chan2022efficient} employ GAN-based 3D generators conditioned on latent vectors to produce category-specific textured 3D assets. Recently, as seen in GigaGAN~\cite{kang2023scaling}, the Generative Adversarial Networks (GANs) methods are better suited for high-frequency details than diffusion models. Furthermore, IT3D~\cite{chen2024it3d} proposes a novel Diffusion-GAN dual training strategy to overcome the view inconsistency challenges. However, the training process of GAN is prone to the issue of mode collapse, which limits the diversity of generation results.

\begin{figure*}[t]
\centering
\includegraphics[width=1\textwidth]{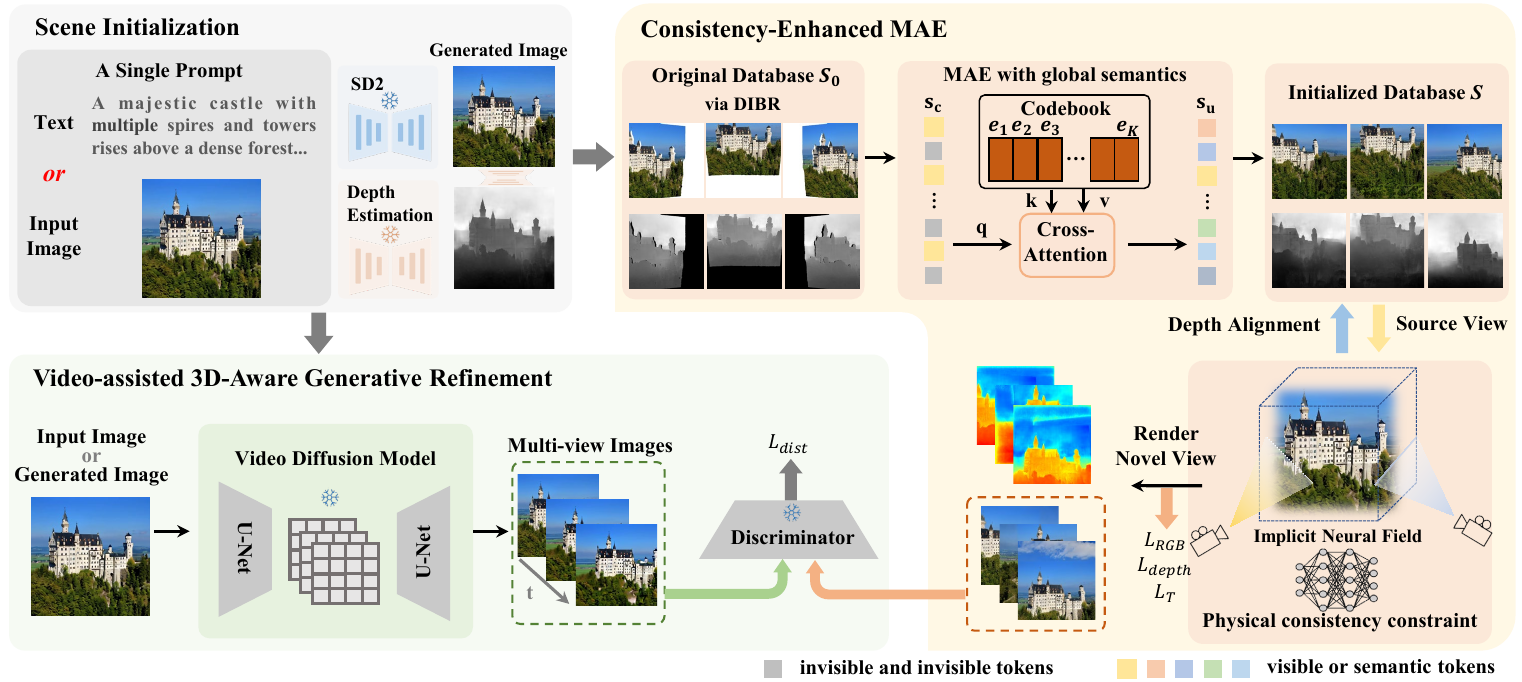} 
\vspace{-3mm}
  \caption{Scene123's pipeline includes two key modules: the consistency-enhanced MAE and the 3D-aware generative refinement module. The former generates adjacent views from an input image via warping, using the MAE model to inpaint unseen areas with global semantics and optimizing an implicit neural field for viewpoint consistency. The latter generates realistic videos from the input image with a pre-trained video generation model, enhancing realism through adversarial loss with rendered images.}
  \vspace{-3mm}
\label{figure: Pipeline}
\end{figure*}

\section{Methodology}

\subsection{Overview}
\vspace{-1mm}
In this paper, we propose a novel 3D scene generation framework based on one prompt, a single image or textual description, ensuring viewpoint consistency and realistic surface textures for both real and synthetically styled scenes, as shown in Fig.~\ref{figure: Pipeline}. We first design the Consistency-Enchanced MAE module to fill unseen areas in novel views by injecting global semantic information. With support views in the initialized database $\mathbf{S}$ generated by the Consistency-Enhanced MAE module, we employ a NeRF network to represent the 3D scene as the physical constraints of view consistency. Furthermore, to enhance the scene's detail and realism, we employ the latest video generation technology, generating high-quality scene videos based on input images and performing adversarial enhancements with the rendered images. Finally, the optimization and implementation details of the model are detailed.


\subsection{Consistency-Enhanced MAE Scene Completion}
\label{sec: Consistency-Enhanced MAE Scene Completion}

\textbf{Scene Initialization.} Given the reference image $\mathbf{I_0}$, notably, for only text prompt input, we utilize the stable diffusion model~\cite{rombach2022high} to generate initial image $\mathbf{I_0}$, we then feed this image into the off-the-shelf depth estimation model~\cite{miangoleh2021boosting}, and take the output as a geometric prior for the target scene, denoted as $\mathbf{D_0}$. Inspired by \cite{zhang2024text2nerf}, we construct an original database $\mathcal{S}_{0}=\left\{\left(\mathbf{I}_{i}, \mathbf{D}_{i}\right)\right\}_{i=1}^{N}$ via the depth image-based rendering (DIBR) method~\cite{fehn2004depth}, where $N$ denotes the number of initial viewpoints. Specifically, for each pixel $x$ in $\mathbf{I_0}$ and its depth value $y$ in $\mathbf{D_0}$, we  compute its corresponding pixel $x_{0 \rightarrow m}$ and depth $y_{0 \rightarrow m}$ on a surrounding view $m$, $\left[x_{0 \rightarrow m}, y_{0 \rightarrow m}\right]^{T}=\mathbf{K} \mathbf{P}_{m} \mathbf{P}_{0}^{-1} \mathbf{K}^{-1}[x, y]^{T}$, where $\mathbf{K}$ and $\mathbf{P}_{m}$ indicate the intrinsic matrix and the camera pose in view $m$. This database provides additional views and depth information, which could prevent the model from overfitting to the initial view. 


\noindent \textbf{MAE Scene Completion.} 
However, the original database $\mathcal{S}_{0}$ will inevitably have missing content since the information in the initial scene is derived from the single image $\mathbf{I_0}$ to construct the initialized database $\mathcal{S}$. Directly applying the original database to the 3D scene representation would inevitably suffer from limited 3D consistency of the generated scenes~\cite{zhang2024text2nerf, hollein2023text2room}. Inspired by ~\cite{he2022masked, hu2023robust, zhang20243d}, we design a Consistency-Enhanced MAE module to effectively ensure the consistency of unseen areas across views. Specifically, we first design a discrete codebook to distribute global information to every invisible area in the original database. We represent the codebook as $\mathcal{E} = \{e_1, e_2, ..., e_{N} \} \in \mathbb{R}^{N \times n_q} $ , where $N$ stands for the total count of prototype vectors, $n_q$ denotes the dimensionality of individual vectors, and $e_i$ symbolizes each specific embedding vector.  Specifically, given the input image $x \in \mathbb{R}^{H \times W \times 3}$, the VQ-VAE~\cite{van2017neural} employs an encoder $E$ to extract a continuous feature representation: $\hat{z}=E(x) \in \mathbb{R}^{h \times w \times c}$, where $h$ and $w$ are the height and width of the feature map. This continuous feature map $\hat{z}$ is then subjected to a quantization process $Q$, aligning it with its nearest codebook entry $e_k$ to obtain its discrete representation $z_q$ as follows: 
\begin{equation}
z_{q}= Q(\hat{z}) := \underset{e_{k} \in \mathcal{E}}{\operatorname{arg min}}\left\|\hat{z}_{ij}-e_{k}\right\|_{2},
\label{equ:total_loss}
\end{equation}
\noindent where $\hat{z}_{ij} \in \mathbb{R}^{c}$. We pre-trained VQVAE on the ImageNet~\cite{russakovsky2015imagenet} dataset to equip the codebook with a more representative and generalized feature set. Subsequently, we fine-tuned it on specific scenes to better capture and represent the unique characteristics of each scene. Moreover, we then utilize the MAE encoder to encode the input images from the original database $\mathcal{S}_{0}=\left\{\left(\mathbf{I}_{i}, \mathbf{D}_{i}\right)\right\}_{i=1}^{N}$, forming the image-conditional $\mathbf{s_c}=\{s_1, s_2, ..., s_M\}$, each embedding vector $s_i$ queries the valuable prior information from the given codebook via the cross-attention mechanism,
\begin{equation}
\begin{split}
\mathrm{Q} \gets f_\mathrm{q}(\mathbf{s_c}),  \qquad\enspace 
\mathrm{K} \gets f_\mathrm{k}(\mathcal{E}),  \qquad\enspace
\mathrm{V} \gets f_\mathrm{v}(\mathcal{E})      \\
\mathbf{s_u} \gets \text{Cross-Attention}(\mathrm{Q},\mathrm{K},\mathrm{V}) = \text{Softmax}(\frac{\mathrm{Q}\mathrm{K}^\mathrm{T}}{\sqrt{d_k}}),
\end{split}
\label{equ:2}
\end{equation}
\noindent where $f_q$, $f_k$, and $f_v$ are the query, key, and value linear projections, respectively. Consequently, the global semantic information contained in the codebook is maintained to distribute global information to every invisible area. Then, we utilize the MAE decoder to decode the feature $\mathbf{s_u}$, deriving the initialized database $\mathcal{S}$, as shown in Fig.~\ref{figure: Pipeline}. Using global semantics as Key and Value in cross-attention allows the model to integrate comprehensive scene information, maintaining coherence. This approach enhances multi-view consistency by providing a unified understanding of the scene, reducing artifacts, and ensuring seamless integration of novel views.

\subsection{3D Scene Representation} 
\label{sec:3D Scene Representation}


With these support views in the initialized database $\mathcal{S}$, along with the initial view $\mathbf{I_0}$, we aim to generate 3D scenes through robust 3D scene representations to provide physical-level surface consistency constraints. In this work, we employ a NeRF network $f_{\theta}$ to represent the 3D scene. Specifically, in NeRF representation, volume rendering~\cite{mildenhall2021nerf} is used to accumulate the color in the radiance fields, 
\begin{equation} 
\mathbf{C}(\mathbf{r})=\int_{t_{n}}^{t_{f}} T(t) \sigma(\mathbf{r}(t)) \mathbf{c}(\mathbf{r}(t), \mathbf{d}) \mathrm{d}t , 
\end{equation}

\noindent where $\mathbf{r}(t)=\mathbf{o}+t \mathbf{d}$ represents the 3D coordinates of sampled points on the camera ray emitted from the camera center $\mathbf{o}$ with the direction $\mathbf{d}$. $t_{n}$ and $t_{f}$ indicate the near and far sampling bounds. $(\mathbf{c}, \sigma)=f_{\theta}(\mathbf{r}(t))$ are the predicted color and density of the sampled point along the ray. 
\begin{equation}
T(t)=\exp \left(-\int_{t_{n}}^{t} \sigma(\mathbf{r}(s)) \mathrm{d} s\right),
\end{equation}
\noindent where $T(t)$ is the accumulated transmittance. Different from NeRF that takes both the 3D coordinate $\mathbf{r}(t)$ and view direction $\mathbf{d}$ to predict the radiance $\mathbf{c}(\mathbf{r}(t), \mathbf{d})$, we omit $\mathbf{d}$ to avoid the effect of view-dependent specularity. 

However, due to the lack of geometric constraint during the depth estimation, the predicted depth values could be misaligned in the overlapping regions~\cite{luo2020consistent}. Despite the design of the consistency-enhanced MAE module greatly improving the inconsistency between different views, the estimated depth rendered from NeRF still may be inconsistent between views. Inspired by Text2NeRF~\cite{zhang2024text2nerf}, we globally align these two depth maps by compensating for mean scale and value differences. Specifically, we first we first perform global alignment by calculating the average $s$ and depth offset $\delta$ to approximate the mean scale and value differences, and then we finetune a pre-trained depth alignment network to produce a locally aligned depth map. More details will be shown in the Appendix.

\begin{figure*}[htbp]
  \centering
  \includegraphics[width=1\linewidth]{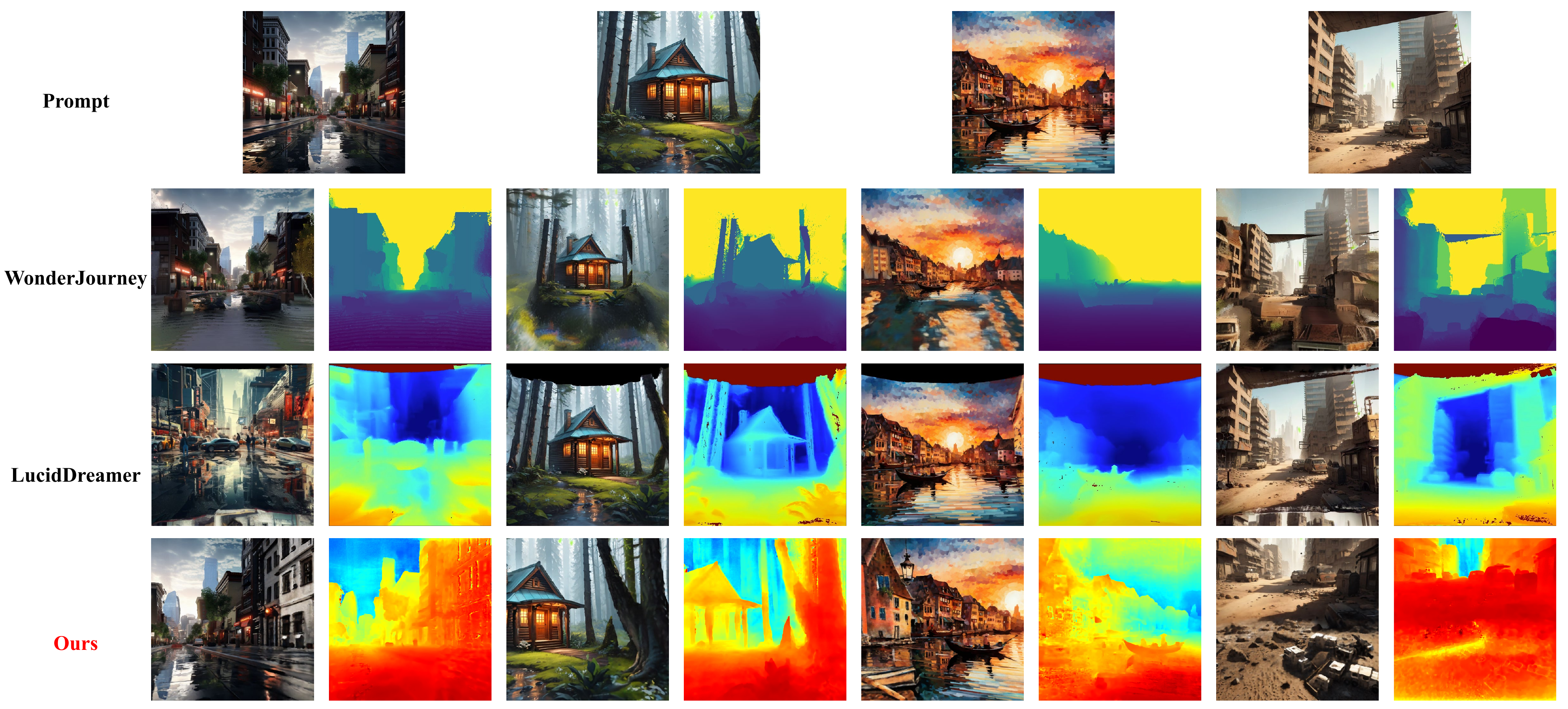}
  \vspace{-3mm}
  \caption{Qualitative results (zoom-in to view better) of methods capable of processing a single image prompt. We both visualize the texture and depth from novel views within the scene. }
  \label{fig: image&text Qualitative results}
  \vspace{-5mm}
\end{figure*}

\subsection{Video-assisted 3D-Aware Generative Refinement}
\label{sec: Video-assisted 3D-Aware Generative Refinement}

\textbf{Support Set Generation.} Given the Reference image $\mathbf{I_0}$, we employ a image-to-video pipeline to generate a support set of enhanced quality, termed $D$, which is conditioned on the input reference image. For the image-to-video pipeline, we opt for Stable Video Diffusion(SVD)~\cite{blattmann2023stable}, which is trained to generate smooth and consistent videos on large-scale datasets of real and high-quality videos. The exposure to superior data quantity and quality makes it more generalizable and multi-view consistent, and the flexibility of the SVD architecture makes it amenable to be finetuned for camera controllability. In the context of the SVD image-to-video~\cite{blattmann2023stable} pipeline, a noise-augmented~\cite{ho2022cascaded} version of the conditioning frame channel-wise is concatenated to the input of the UNet~\cite{ronneberger2015u}. In addition, the temporal attention layers in SVD naturally assist in the consistent multi-view generation without needing any explicit 3D structures like in~\cite{liu2023syncdreamer}. 

\begin{figure}[htbp]
  \centering
  \includegraphics[width=1\linewidth]{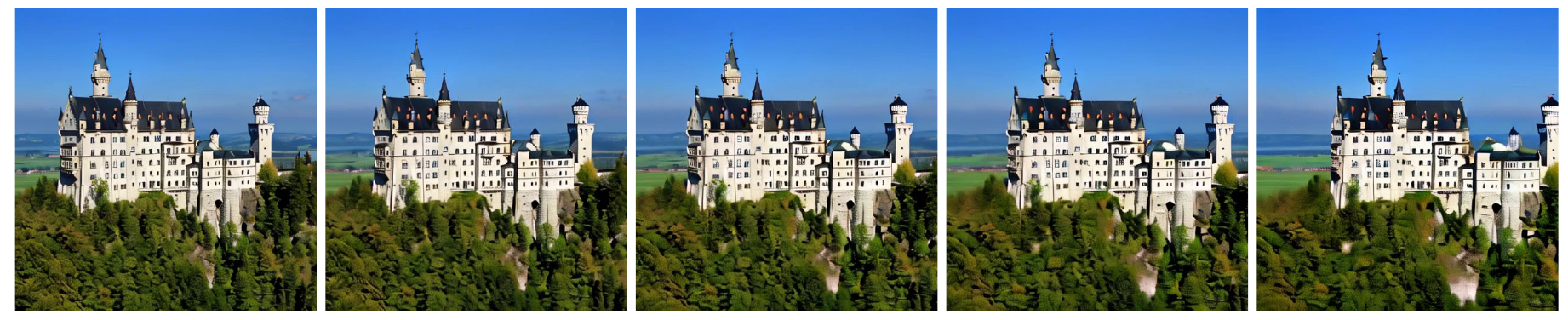}
    \vspace{-6mm}
  \caption{Data samples generated via image-to-video model.}
  \label{fig: data samples}
\end{figure}

\noindent \textbf{3D-Aware Generative Refinement.} The capabilities of Generative Adversarial Networks (GANs) shine in scenarios involving datasets characterized by high variance~\cite{chan2022efficient}. GANs have the ability to learn both geometry and texture-related knowledge from datasets, subsequently guiding the model to converge towards the same high-quality distribution exhibited by the generated support set. In our approach, we designate the video diffusion model as a generator. As shown in Fig.~\ref{fig: data samples}, given a reference image, SVD can generate consistent multi-view images at one time, constructing the generated support set. We then incorporate a discriminator initialized with random values. In this setup, the Support Set $D$ is treated as real data, while the renderings of the 3D neural radiance field model represent fake data. The role of the discriminator involves learning the distribution discrepancy between the renderings and $D$, subsequently contributing to the discrimination loss, which in turn updates the 3D neural radiance field model. This 3D-aware generative refinement module utilizes the discrimination loss $L_{\mathbf{dist}}$, 
 which can help guide the updating direction and enhance the model’s ability to produce intricate geometry and texture details.

The camera motion in the SVD model is sometimes limited\cite{blattmann2023stable}, making it unsuitable for directly training the NeRF model. Instead, we use our 3D-Aware Generative Refinement approach, which leverages a discriminator to optimize the process. Despite the small range of camera motion in the images generated by SVD, the adversarial training process provides necessary regularization, demonstrating that even with limited camera motion, the discrimination loss remains effective and contributes to generating high-quality 3D scenes.

\begin{figure*}[htbp]
  \centering
  \vspace{-3mm}
  \includegraphics[width=1\linewidth]{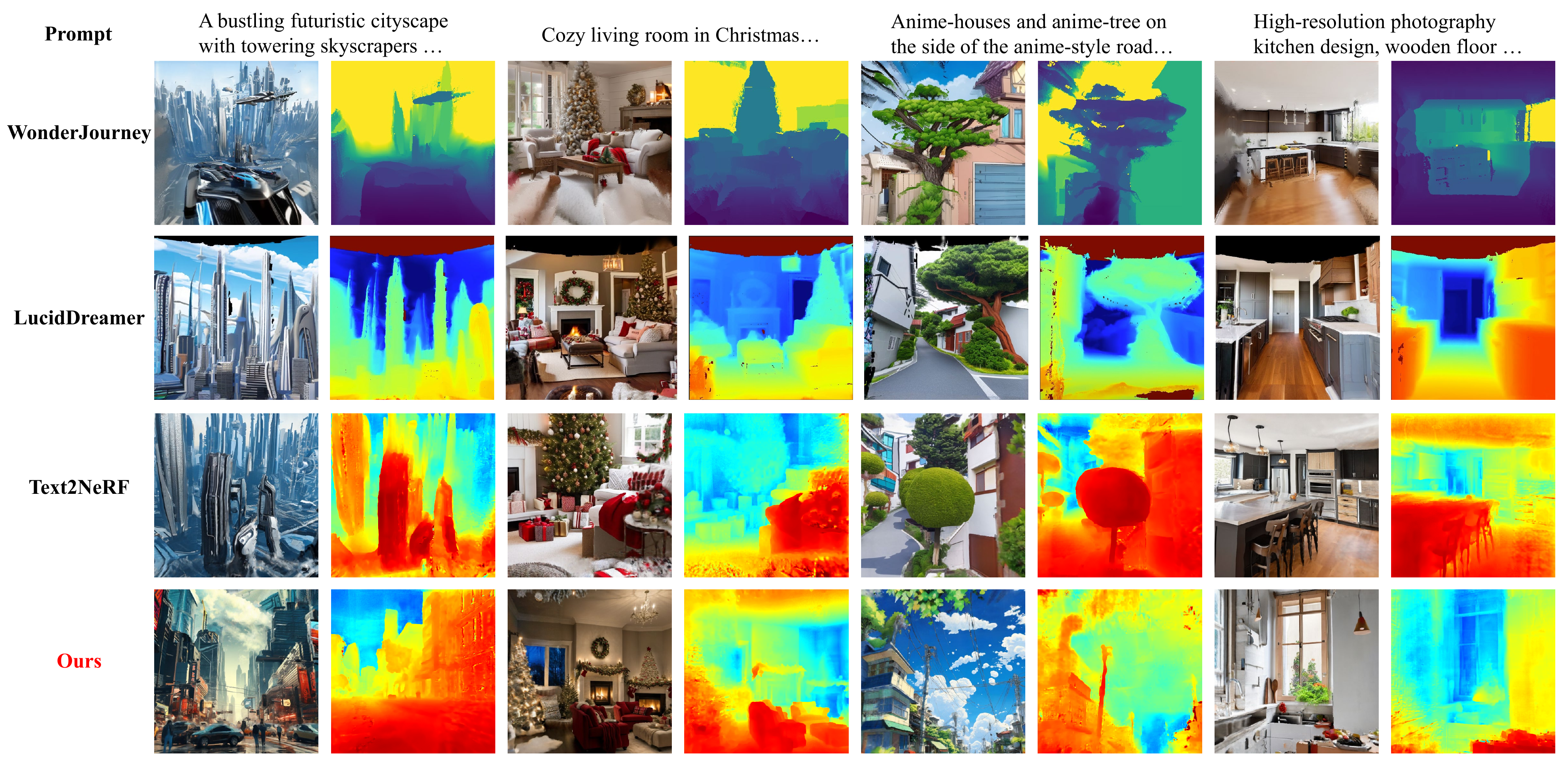}
  \caption{Qualitative results (zoom-in to view better) of methods that generate scenes from textual input. We both visualize the texture and depth from novel views within the scene. }
  \label{fig: text-driven Qualitative results}
  \vspace{-5mm}
\end{figure*}

\subsection{Optimization and Implementation Details} 
\label{sec: Optimization}
 In addition to discrimination loss, we also utilize a $L_{2}$ loss, depth loss, and a transmittance loss to optimize the radiance field of the 3D scene, following previous NeRF-based works~\cite{chen2022tensorf,song2023nerfplayer,zhang2024text2nerf}. The RGB loss $L_{\mathbf{RGB}}$ is defined as a $L_{2}$ loss between the render pixel $\boldsymbol{C}^{R}$ and the color $\boldsymbol{C}$ generated by the MAE model. Different from previous works that employ regularized depth losses to handle uncertainty or scale-variant problems~\cite{roessle2022dense,sargent2023vq3d}, we adopt a stricter depth loss $L_{\mathbf{depth}}$ to minimize the $L_{2}$ distance between the rendered depth and the estimated depth. Moreover, we compute a depth-aware transmittance loss $L_{\mathbf{T}}$~\cite{jain2022zero,zhang2024text2nerf} to encourage the NeRF network to produce empty density before the camera ray reaches the expected depth $\mathbf{\hat{z}}$,   $L_\mathbf{T}=\|\mathbf{T}(t) \cdot \mathbf{m}(t)\|_{2}$, where $\mathbf{m}(t)$ is a mask indicator that satisfies $\mathbf{m}(t)=1$ when $t < \mathbf{\hat{z}}$, otherwise $\mathbf{m}(t)= 0$. $\mathbf{\hat{z}}$ is the pixel-wise depth value in the estimated depth map, and $\mathbf{T}(t)$ is the accumulated transmittance. Therefore, the total loss function is then defined as,

\begin{equation}
L_{\mathbf{total}}=L_{\mathbf{RGB}}+ \lambda_{d} L_{\mathbf{depth}} + \lambda_{t} L_{\mathbf{T}} + \lambda_{dist} L_{\mathbf{dist}},
\end{equation}
\noindent where $\lambda_{d}$, $\lambda_{T}$, $\lambda_{dist}$ are constant hyperparameters 
balancing depth, transmittance, and discrimination losses.

\noindent \textbf{Implementation Details.} We implement our Scene123 with the Pytorch framework~\cite{paszke2019pytorch} and adopt TensoRF~\cite{chen2022tensorf} as the radiance field. To ensure TensoRF can accommodate scene generation over a large view range, we position the camera near the center of the NeRF bounding box and configure it with outward-facing viewpoints. The dimension of the masked codebook is 2048$\times$16. For only text prompt input, we utilize the stable diffusion model in version 2.0~\cite{rombach2022high} to generated initial image $\mathbf{I_0}$ related to the input prompt. Moreover, for depth estimation, we use the boosting monocular depth estimation method~\cite{miangoleh2021boosting} with pre-trained LeReS model~\cite{yin2021learning} to estimate the depth for each view. For the image-to-video pipeline, we opt for the Stable Video Diffusion in version SVD-XT, which is the same architecture as SVD~\cite{blattmann2023stable} but finetuned for 25 frame generation. During training, we use the same setting as~\cite{chen2022tensorf} for the optimizer and learning rate and set the hyperparameters in our objective function as $\lambda_d = 0.005$, $\lambda_t = 0.001$, $\lambda_{dist} = 0.001$.

\section{Experiments}
\label{Experiments}

\subsection{Experimental Setup}


\textbf{Dataset and baselines.} Since the perpetual 3D scene generation is a new task without an existing dataset, we use real or generated high-quality images as input~\cite{yu2023wonderjourney} for evaluation in our experiments. We consider two state-of-the-art 3D scene generation methods as our baselines, WonderJourney~\cite{yu2023wonderjourney} and LucidDreamer~\cite{chung2023luciddreamer} to compare the performance of the image or text prompt as a condition input for 3D scene generation. WonderJourney and LucidDreamer rely on an off-the-shelf general-purpose depth estimation model to project the hallucinated 2D scene extensions into a 3D representation. Specifically, WonderJourney designs a fully modularized model to generate sequences of 3D scenes. LucidDreamer utilizes Stable Diffusion~\cite{rombach2022high} and 3D Gaussian splatting~\cite{kerbl20233d} to create diverse high-quality 3D scenes. For text prompts, 
besides WonderJourney and LucidDreamer, we include 
Text2NeRF~\cite{zhang2024text2nerf} as the baseline, which performs well for the text-to-3D generation. Text2NeRF generates NeRFs from text with the aid of text-to-image diffusion models to represent 3D scenes.

\begin{table*}[hbtp]
  \centering
  \resizebox{0.72\linewidth}{!}{
    \scalebox{1}{\begin{tabular}{cccccc}
    \toprule
    \multicolumn{1}{c}{\multirow{2}[1]{*}{Method}} & \multicolumn{1}{c}{\multirow{2}[1]{*}{Input}}  & \multicolumn{3}{c}{Visual Quality} & \multicolumn{1}{c}{\multirow{2}[1]{*}{Optimization Time (hours)}}  \\
        & & CLIP-Similarity↑ & BRISQUE↓ & NIQE↓ &   \\
    \cmidrule(r){1-1}  \cmidrule(r){2-2}  \cmidrule(r){3-5} \cmidrule(r){6-6}  
    WonderJourney~\cite{yu2023wonderjourney}  & Image\&Text   & 27.480 & 67.012 & 12.022 &  \textbf{0.208}   \\ 
    LucidDreamer~\cite{chung2023luciddreamer}  & Image\&Text   & 26.663 & 46.266 & 6.652 &  0.220  \\ 
    Text2NeRF~\cite{zhang2024text2nerf}  & Text   & 28.695 & 24.498 & 4.618 &  1.525  \\ 
    \textbf{Scene123 (Ours)}  & Image\&Text  & \textbf{30.544} & \textbf{20.324} & \textbf{2.522} &  1.433   \\ 
    \bottomrule
    \end{tabular}
    }
    }
    \vspace{-2mm}
   \caption{Quantitative comparison results of our method with the baseline WonderJourney, LucidDreamer and Text2NeRF ↑ means the higher, the better, ↓ means the lower, the better. }
   \vspace{-2mm}
  \label{tab: The image Quantitative result}%
\end{table*}%

\noindent \textbf{Evaluation Metrics.} Following Text2NeRF, we evaluate the quality of our generated images using CLIP Score (CLIP-Similarity), Blind/Referenceless Image Spatial Quality Evaluator (BRISQUE)~\cite{mittal2012no} and Natural Image Quality Evaluator
(NQIE)~\cite{mittal2012making}.

\subsection{Performance Comparisons} As shown in Tab.~\ref{tab: The image Quantitative result}, we evaluate the quality of generated 3D scenes across baselines quantitatively and report the average evaluation scores of CLIP-Similarity BRISQUE and NIQE for the generated images produced by different methods. Clearly, our method surpasses the baselines by generating higher-quality 3D scenes, as indicated by lower BRISQUE and NIQE values. Moreover, our method ensures the semantic relevance between the generated scene and the input text, resulting in a higher CLIP score.
The qualitative results are drawn in Fig.~\ref{fig: image&text Qualitative results} and Fig.~\ref{fig: text-driven Qualitative results}. Our method can ensure more viewpoint consistency and realistic surface textures for both real and synthetically styled scenes, giving a single image or textual description. Obviously, Fig.~\ref{fig: image&text Qualitative results}, given a reference image, WonderJourney and LucidDreamer seem to generate 3D scenes with artifacts, while our method can generate more semantic relevant to the given image, maintaining the multi-view consistency. In  Fig.~\ref{fig: text-driven Qualitative results}, with textual prompt input, WonderJourney can inevitably generate artifacts. While LucidDreamer and Text2NeRF are likely to produce distorted or inconsistent viewpoint images. Consequently, our method demonstrates superior qualitative performance compared to these baseline approaches.

\subsection{Ablation Studies and Analysis} 

\begin{figure}[htbp]
  \centering
  \includegraphics[width=0.9\linewidth]{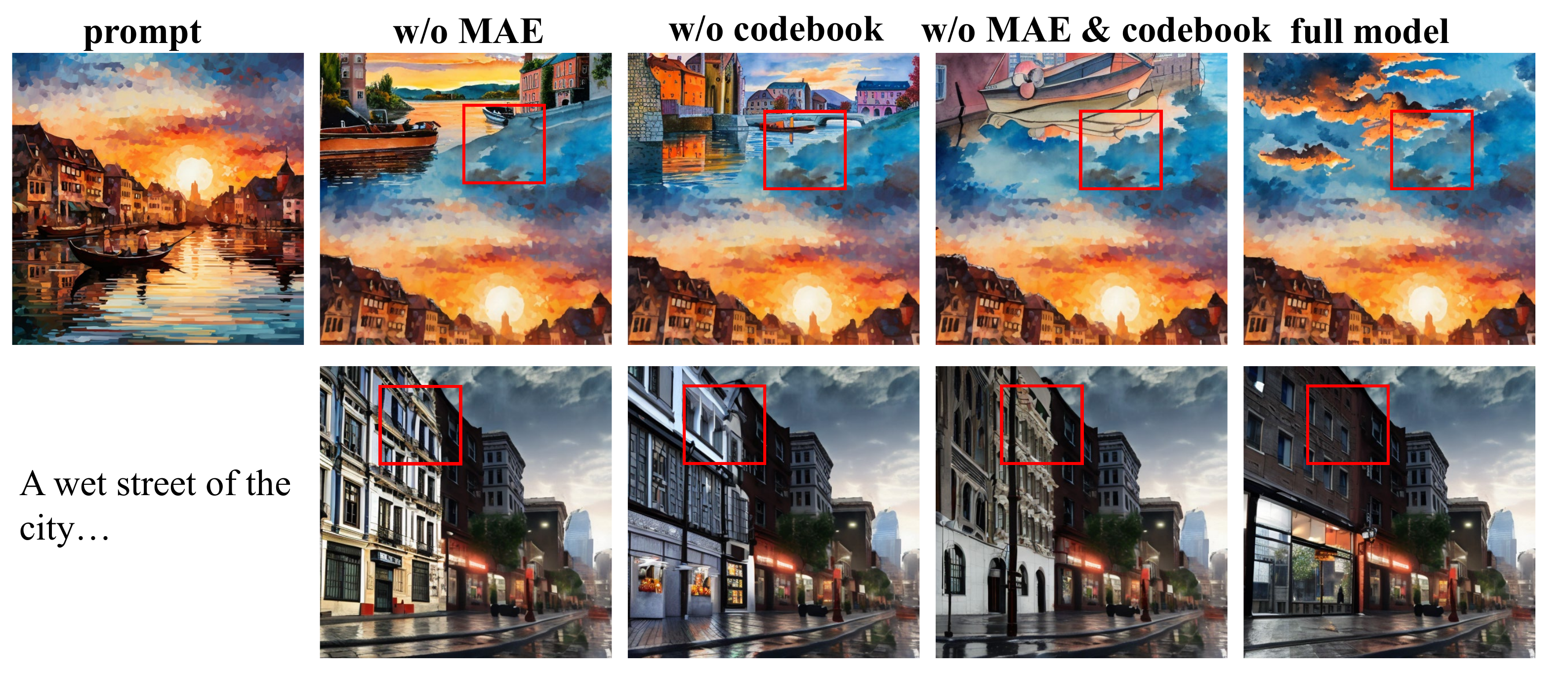}
  \vspace{-2mm}
  \caption{ Results of the effectiveness of the MAE module and the codebook.}
 \label{fig :Experiment results of the effectiveness of the Consistency-Enhanced MAE.}
\end{figure}

\textbf{Effectiveness of the Consistency-Enhanced MAE.} To verify the effectiveness of the Consistency-Enhanced MAE, we conduct ablation studies on different strategies: removing the MAE scene completion, denoted as w/o MAE; removing the masked VQ-VQE codebook, denoted w/o codebook; removing both MAE scene completion and the masked VQ-VAE codebook, denoted as w/o MAE$\&$codebook. For w/o MAE, we utilize the stable diffusion (sd) inpainting to complete the scene, which is utilized in LucidDreamer. As shown in Tab.~\ref{tab:Experiment results of the effectiveness of the Consistency-Enhanced MAE.} and Fig.~\ref{fig :Experiment results of the effectiveness of the Consistency-Enhanced MAE.}, the integration of the consistency-enhanced MAE and the codebook significantly contributes to the generation of coherent 3D scenes.  Fig.~\ref{fig :Experiment results of the effectiveness of the Consistency-Enhanced MAE.} demonstrates visually that when the model removes either of the MAE and codebook modules, it causes inconsistencies between the different views.  This superiority of our  Consistency-Enhanced MAE in handling detailed complementation is evident in Fig.~ \ref{fig :Experiment results of the effectiveness of the Consistency-Enhanced MAE.}.

\begin{table}[hbtp]
  \centering
  \resizebox{0.72\linewidth}{!}{
    \scalebox{1}{\begin{tabular}{cccc}
    \toprule
     Method &  CLIP↑  & BRISQUE↓ & NIQE↓ \\
     \cmidrule(r){1-1} \cmidrule(r){2-4}
    w/o MAE      & 27.745   & 41.243   & 6.024  \\
    w/o codebook     & 27.983   &  38.335   & 5.834  \\
    w/o MAE$\&$codebook      &  26.674   & 44.234   &  6.653  \\
     \cmidrule(r){1-1} \cmidrule(r){2-4}
    \textbf{full model}   & \textbf{30.544}   & \textbf{20.324}   & \textbf{2.522}  \\
    \bottomrule
    \end{tabular}%
    }
    }
    \vspace{-2mm}
   \caption{Ablation experiments regarding the MAE module and the codebook.}
    \vspace{-6mm}
    \label{tab:Experiment results of the effectiveness of the Consistency-Enhanced MAE.}
\end{table}%

\begin{figure}[htbp]
  \centering
  \includegraphics[width=0.9\linewidth]{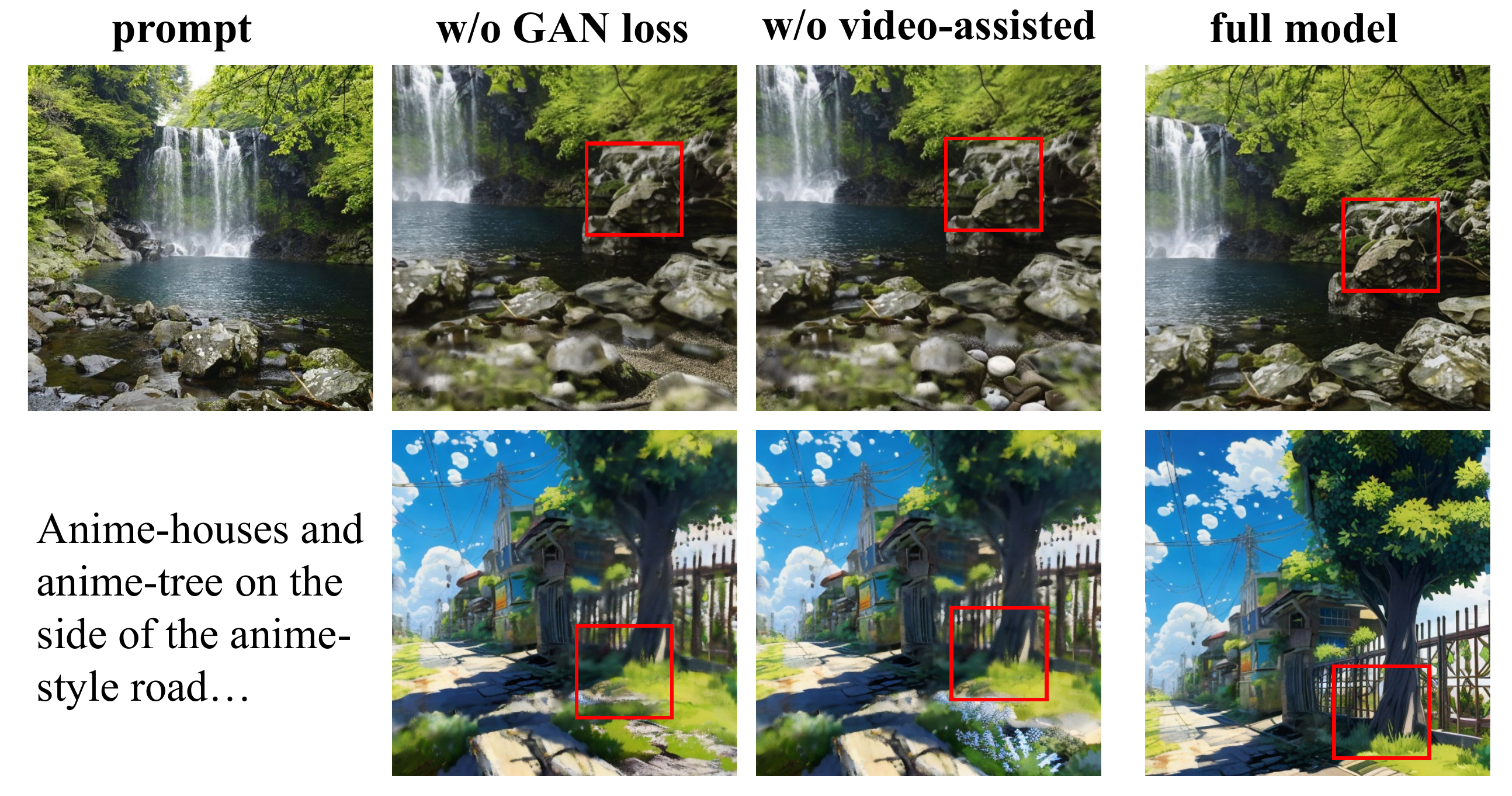}
  \vspace{-2mm}
  \caption{Results of the effectiveness of the  3D-aware generative refinement module.}
 \label{fig :Experiment results of the effectiveness of the GAN loss.}
\end{figure}

\noindent \textbf{Effectiveness of the 3D-aware generative refinement module.} To verify the effectiveness of the video-assisted 3D-aware generative refinement module, we conduct ablation studies on different strategies, including removing the discrimination loss, denoted as w/o GAN loss; replacing the real support set generated by the image-to-video pipeline with a set containing the same number of images using reference image duplication, denoted as w/o video-assisted. As shown in Tab.~\ref{tab:Experiment results of the effectiveness of the GAN loss.} and Fig.~\ref{fig :Experiment results of the effectiveness of the GAN loss.}, the incorporation of the GAN-based training strategy significantly enhances the model's ability to render detailed textures and complex geometries.  The full model achieves the lowest FID values, while similar in other metrics, indicating the 3D-aware generative refinement module plays an important role in providing intricate geometry and texture details.

\begin{table}[hbtp]
  \centering
  \resizebox{0.75\linewidth}{!}{
    \scalebox{1}{\begin{tabular}{cccc}
    \toprule
    Method  & CLIP↑  &  BRISQUE↓ & NIQE↓\\
     \cmidrule(r){1-1} \cmidrule(r){2-4}
    w/o GAN loss      & 25.234   & 33.234   & 7.234  \\
    w/o video-assisted     & 28.341   & 24.342   & 5.342  \\
     \cmidrule(r){1-1} \cmidrule(r){2-4} 
    \textbf{full model}   & \textbf{30.544}   & \textbf{20.324}   & \textbf{2.522} \\
    \bottomrule
    \end{tabular}
    }
    }
    \vspace{-2mm}
   \caption{Ablation experiments regarding the  3D-aware generative refinement module.}
   \vspace{-2mm}
   \label{tab:Experiment results of the effectiveness of the GAN loss.}
\end{table}

\section{Conclusions and limitations}
\label{Conclusions and limitations}

\textbf{Conclusions.} In this paper, we introduce Scene123, which surpasses existing 3D scene generation methods in scene consistency and realism, providing finer geometry and high-fidelity textures.
Our method mainly relies on the consistency-enhanced MAE and the 3D-aware generative refinement module. The former exploits the inherent scene consistency constraints of implicit neural fields and integrates them with the MAE model with global semantics to inpaint adjacent views, ensuring viewpoint consistency. 
The latter uses the video generation model to produce realistic videos, enhancing the detail and realism of rendered views. 
With the help of these two modules, our method can generate high-quality 3D scenes from a single input prompt, whether real, virtual, or object-centered settings.

\noindent \textbf{Limitations.} Our method is both innovative and effective. However, the optimization process remains an area for potential enhancement, as it is currently constrained by our use of implicit neural fields to enforce physical consistency. In our future research, we aim to explore faster, more easily optimized, and sustainable 3D scene representations. These will serve as mechanisms to articulate consistency surface constraints, thus accelerating the generation process.

\clearpage

\footnotesize
\bibliography{aaai25}

\clearpage

\appendix

\maketitle

\section{Appendix}

\subsection{Implementation Details}

 We implement our Scene123 with the Pytorch framework~\cite{paszke2019pytorch} and adopt TensoRF~\cite{chen2022tensorf} as the radiance field. To ensure TensoRF can accommodate scene generation over a large view range, we position the camera near the center of the NeRF bounding box and configure it with outward-facing viewpoints. The dimension of the masked codebook is 2048$\times$16. For only text prompt input, we utilize the stable diffusion model in version 2.0~\cite{rombach2022high} to generated initial image $\mathbf{I_0}$ related to the input prompt. Moreover, for depth estimation, we use the boosting monocular depth estimation method~\cite{miangoleh2021boosting} with pre-trained LeReS model~\cite{yin2021learning} to estimate the depth for each view. For the image-to-video pipeline, we opt for the Stable Video Diffusion in version SVD-XT, which is the same architecture as SVD~\cite{blattmann2023stable} but finetuned for 25 frame generation. During training, we use the same setting as~\cite{chen2022tensorf} for the optimizer and learning rate and set the hyperparameters in our objective function as $\lambda_d = 0.005$, $\lambda_t = 0.001$, $\lambda_{dist} = 0.001$.

\noindent \textbf{GAN-based Training Strategy Details.} For the incorporated discriminator, we adopt a similar architecture, regularization function, and loss weight as the EG3D model~\cite{chan2022efficient}, with some distinctions. The vanilla structure of a 3D GAN involves a 3D generator that incorporates a super-resolution module, accompanied by a discriminator that accepts both coarse and fine image inputs~\cite{chan2022efficient}. Notably, due to the contextual disparities between text-to-3D and 3D GAN applications, we opt to omit the super-resolution component from the 3D GAN architecture. This choice stems from the consistent need to extract mesh or voxel representations from the 3D model within the context of text-to-3D. 

The camera motion in the SVD model is sometimes limited~\cite{blattmann2023stable}, making it unsuitable for directly training the NeRF model.  Instead, we use our 3D-Aware Generative Refinement approach, which leverages a discriminator to optimize the process. In this setup, the video diffusion model's output is treated as real data, while the 3D neural radiance field model's renderings are treated as fake data. Despite the small range of camera motion in the images generated by SVD, the adversarial training process provides necessary regularization. The discriminator learns to distinguish between real support set images and fake renderings, thereby pushing the neural radiance field model to produce outputs that are indistinguishable from real images. This continuous adversarial interaction compels the generator to improve by learning finer details and more accurate textures, ultimately enhancing the quality of the 3D scene generation. This method ensures that even with limited camera motion, the GAN loss remains effective and contributes to generating high-quality 3D scenes, as shown in Fig.~\ref{fig :Experiment results of the effectiveness of the GAN loss.} in the manuscript.

\noindent \textbf{Depth alignment details.} We use the depth estimation model $f_e$ to estimate the depth map $D_k^E$ for the initial view $\mathbf{I_0}$. Note that, unlike the depth map $D_0$ of the initial view, $D_k^E$ cannot be directly taken as the supervision to update the radiance field since it is predicted independently and could conflict with known depth maps such as $D_k^R$ in the overlapping regions. To solve this issue, we implement depth alignment to align the estimated depth map to the known depth values in the radiance field~\cite{zhang2024text2nerf}.
 
Due to the lack of geometric constraint during the depth estimation, the predicted depth values could be misaligned in the overlapping regions \cite{luo2020consistent}, for example, the estimated depth $D_k^E$ of the inpainted view may be inconsistent with the depth $D_k^R$ rendered from NeRF since $D_k^R$ is constrained by previous known views.
The inconsistency is manifested in two aspects: scale difference and value difference. For instance, the \textit{distance difference} of two pixel-aligned spatial points and the \textit{depth value} of a specific point could be both different in depth maps estimated from different views. The former is the scale difference and the latter is the value difference. In the case of scale difference, we cannot align both points by shift processing because even if we align the depth value of one of the points, the other point is still misaligned. To eliminate the scale and value differences between the overlapping regions of the rendered depth map $D_k^R$ and the estimated depth map $D_k^E$ of the novel view, we introduce a two-stage depth alignment strategy. Specifically, we first globally align these two depth maps by compensating for mean scale and value differences. Then we finetune a pre-trained depth alignment network to produce a locally aligned depth map. 

To determine the mean scale and value differences, we first randomly select $M$ pixel pairs from the overlapping regions and deduce their 3D positions under depth $D_k^R$ and $D_k^E$, denoted as $\left \{ (\mathbf{x}_j^R,\mathbf{x}_j^E) \right \}_{j=1}^M$. Next, we calculate the average scaling score $s$ and depth offset $\delta$ to approximate the mean scale and value differences:
\begin{equation}
  s = \frac{1}{M-1}\sum_{j=1}^{M-1}\frac{\Vert\mathbf{x}_j^R-\mathbf{x}_{j+1}^R\Vert_2}{\Vert\mathbf{x}_j^E-\mathbf{x}_{j+1}^E\Vert_2},
  \label{eq:scale}
\end{equation}
\begin{equation}
  \delta = \frac{1}{M}\sum_{j=1}^{M}\left(z\left(\mathbf{x}_j^R\right)-z\left(\mathbf{\hat{x}}_j^E\right)\right),
  \label{eq:offset}
\end{equation}
where $\mathbf{\hat{x}}_j^E = s \cdot \mathbf{x}_j^E$ indicates the scaled point and $z(\mathbf{x})$ represents the depth value of point $\mathbf{x}$. Then $D_k^E$ can be globally aligned with $D_k^R$ by $D_k^{global} = s \cdot D_k^E + \delta$.

Since depth maps used in our pipeline are predicted by a network, the differences between $D_k^R$ and $D_k^E$ are not linear, that is why the global depth aligning process cannot solve the misalignment problem. To further mitigate the local difference between $D_k^{global}$ and $D_k^R$, we train a pixel-to-pixel network $f_{\psi}$ for nonlinear depth alignment. During optimization of each view, we optimize the parameter $\psi$ of the pre-trained depth alignment network $f_{\psi}$ by minimizing their least square error in the overlapping regions:
\begin{equation}
  \mathop{\min}_{\psi} \left\| \left( f_{\psi}(D_k^{global}) - D_k^R \right) \odot M_k \right\|_2.
  \label{eq:local_align}
\end{equation}
Finally, we can derive the locally aligned depth using the optimized depth alignment network: $\hat{D}_k = f_{\psi}(D_k^{global})$.

\begin{figure*}[htbp]
  \centering
\includegraphics[width=1\linewidth]{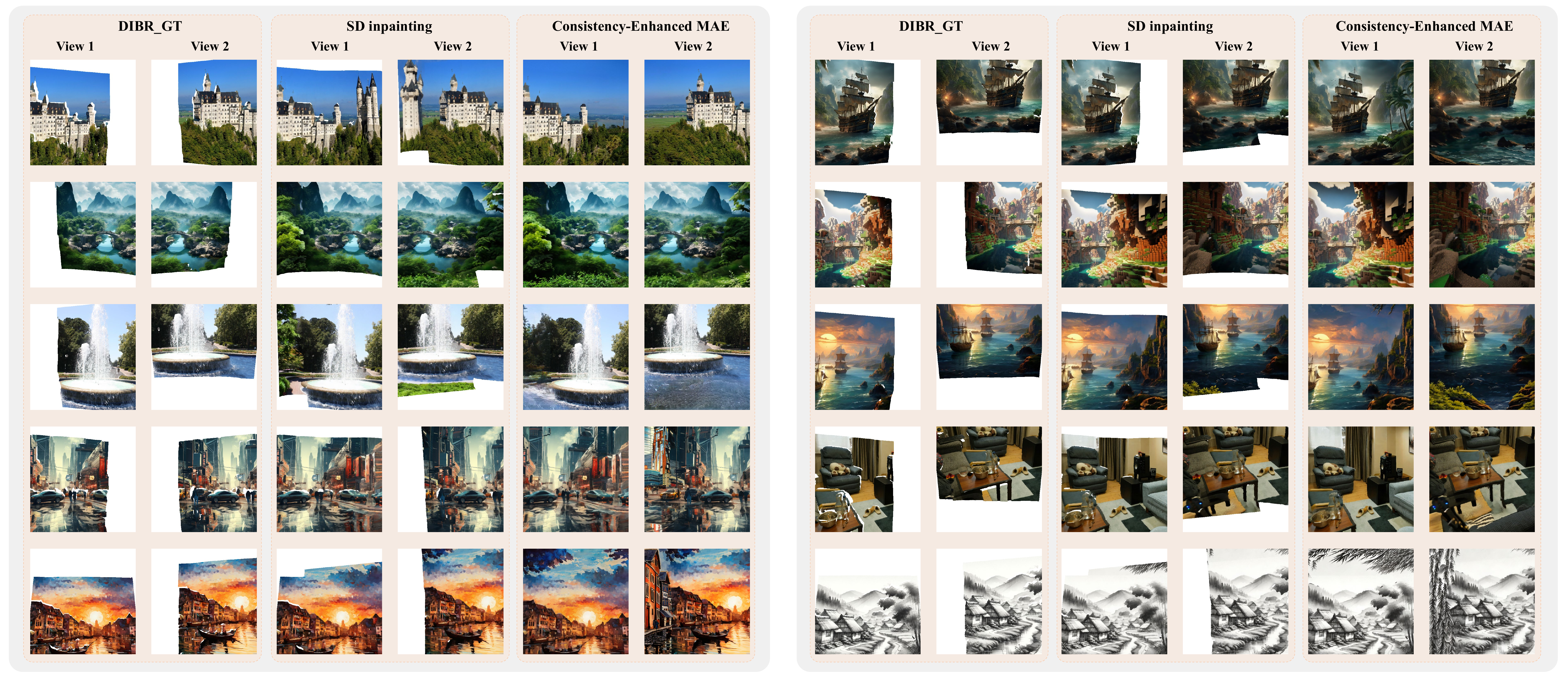}
  \caption{Comparisons of SD inpainting and our Consistency-Enhanced MAE in handling detailed complementation.}
  \label{fig:supp3}
\end{figure*}

\subsection{Additional analysis}

\noindent \textbf{Comparison between SD inpainting and Our Consistency-Enhanced MAE in detail completion.} Our Consistency-Enhanced MAE ensures consistency across views by using a learnable codebook to distribute global information to every invisible area. Traditional VAE methods quantize the latent space, leading to artifacts at the boundaries between quantized latents. In contrast,  the cross attention mechnism allows for dynamic and flexible feature weighting, better captures global context, and preserves detailed information without the loss associated with quantization. The learnable codebook continuously adapts during training, enhancing the model's stability and performance in generating high-quality 3D scenes.

However, the stable diffusion (SD) model,  which is originally utilized to inpaint images in LucidDreamer~\cite{chung2023luciddreamer}, is less effective for the multi-view complementation task. While SD can achieve complementation through conditional diffusion, it struggles with complex structures and view-consistency tasks, as it is not tailored for this purpose. Although SD performs well in generating new images, it may not be as effective in complementing local details compared to MAE, which is specifically designed for this task. This superiority of our  Consistency-Enhanced MAE in handling detailed complementation is evident in Fig.~\ref{fig:supp3}.

\begin{figure}[t]
  \centering
  \includegraphics[width=1\linewidth]{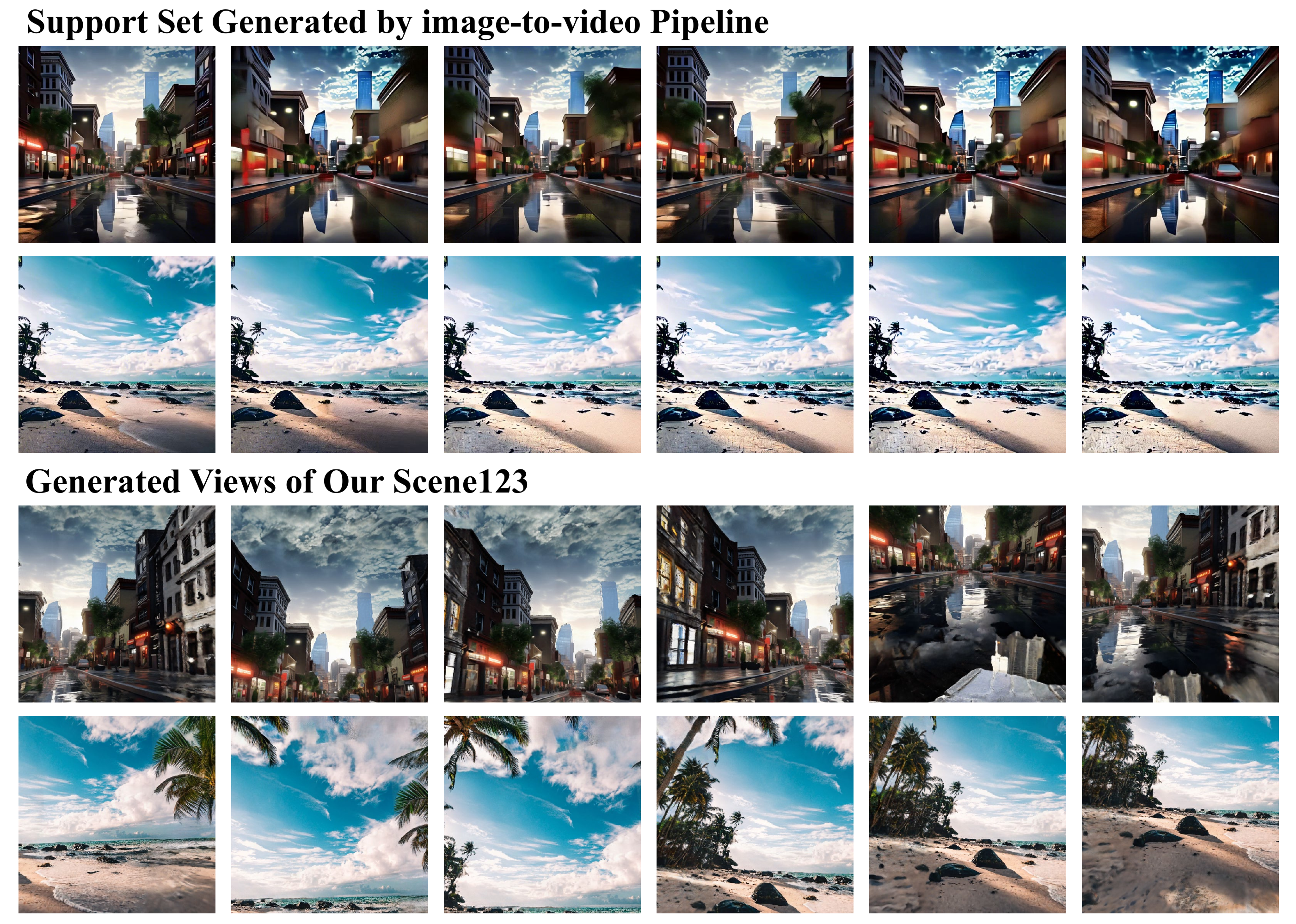}
  \caption{Comparisons between support set generated by image-to-video pipeline and the generated views of Our Scene123.}
  \vspace{-4mm}
  \label{fig:supp2}
\end{figure}

\noindent \textbf{Video Generation Model promotes our Scene123's quality.} The 3D scenes generated through our 3D-Aware Generative Refinement module are of higher quality than the image-to-video generation pipeline. As depicted in Fig.~\ref{fig:supp2}, our Scene123 does not strictly require a high-quality support set. The novel views generated by our method are generally superior to those produced by the image-to-video pipeline. This is because our 3D-Aware Generative Refinement uses a discriminator to manage optimization, distinguishing between real and fake images. This adversarial interaction improves the generator's ability to learn finer details and more accurate textures, enhancing the overall quality of 3D scene generation. This demonstrates our approach's adaptability in handling image-to-video generation failures.

\subsection{User Study}

For completeness, we follow previous works~\cite{wang2024prolificdreamer,yu2023wonderjourney} and conduct a user study by comparing Scene123 with WonderJourney~\cite{yu2023wonderjourney}, LucidDreamer~\cite{chung2023luciddreamer} under 40 image prompts, 20 image prompts for each baseline. For text prompt input, we also compare Scene123 with WonderJourney~\cite{yu2023wonderjourney}, LucidDreamer~\cite{chung2023luciddreamer} and Text2NeRF~\cite{zhang2024text2nerf} under 60 text prompts, 20 text prompts for each baseline. The participants are shown the generated results of our Scene123 and baselines and asked to choose the better one in terms of fidelity, details and vividness. We collect results from 39 participants, yielding 3120 pairwise comparisons. The results are shown in Tab.~\ref{tab: User Study.}. Our method outperforms all of the baselines.

\begin{table}[htbp]
  \centering
  \resizebox{1\linewidth}{!}{
    \scalebox{1}{\begin{tabular}{cccc}
    \toprule
    Method & WonderJourney & LucidDreamer & Text2NeRF  \\
    \cmidrule(r){1-1} \cmidrule(r){2-4} 
    Prefer baseline & 31.12 & 23.42 & 46.27  \\
    Prefer Scene123 (Ours)  & \textbf{68.88}   & \textbf{76.58}   & \textbf{53.73}    \\
    \bottomrule
    \end{tabular}%
    }
    }    \caption{Results of user study. The percentage of user preference (↑) is reported in the table.}
  \label{tab: User Study.}%
\end{table}%

\subsection{More Qualitative Results}

\begin{figure*}[hbtp]
  \centering
  \includegraphics[width=1\linewidth]{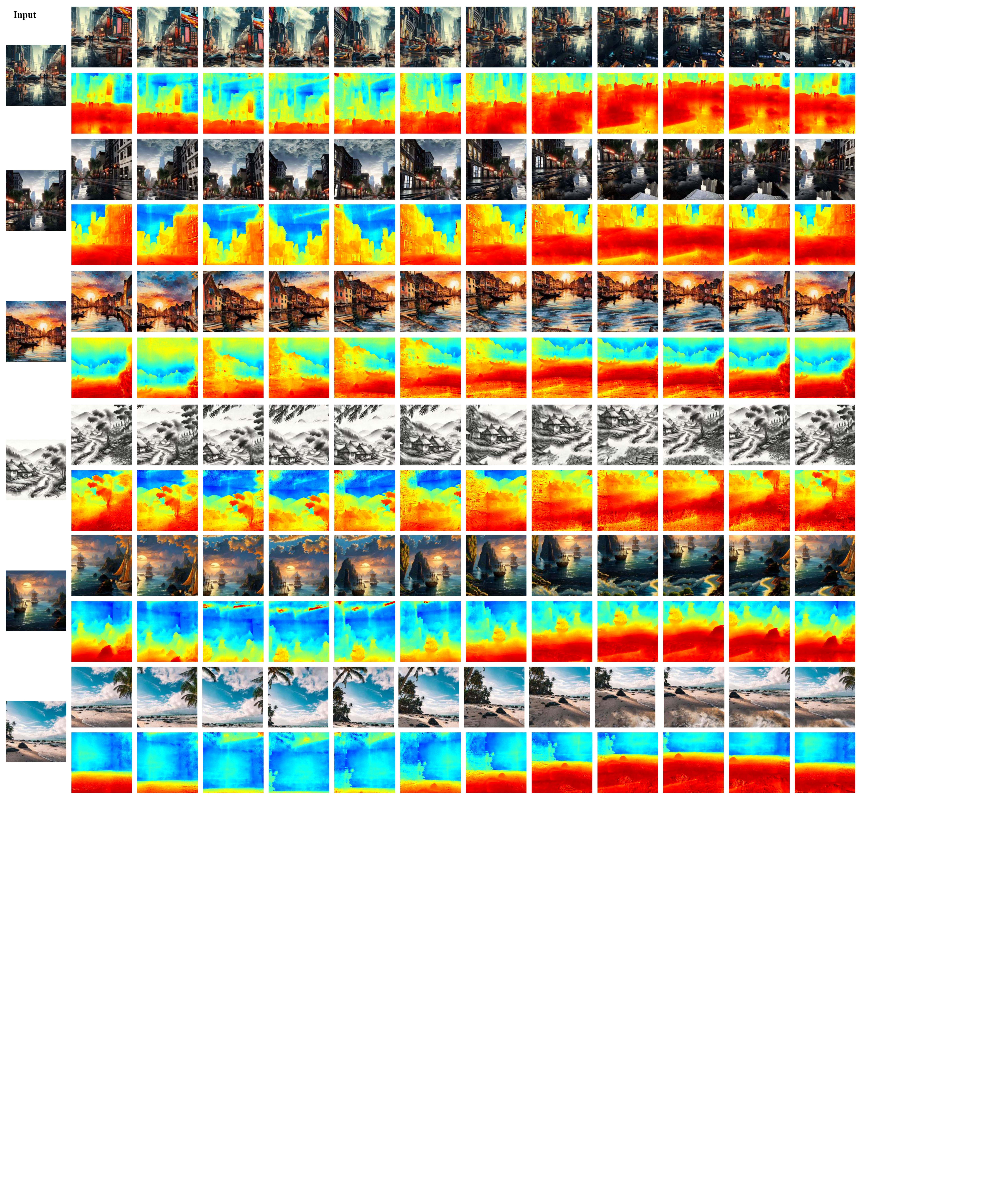}
   Part 1\,/\,4
  \caption{More qualitative examples generated by our Scene123 from a single image input.}
  \label{fig:supp_result1}
\end{figure*}

\begin{figure*}[hbtp]
  \centering
  \includegraphics[width=1\linewidth]{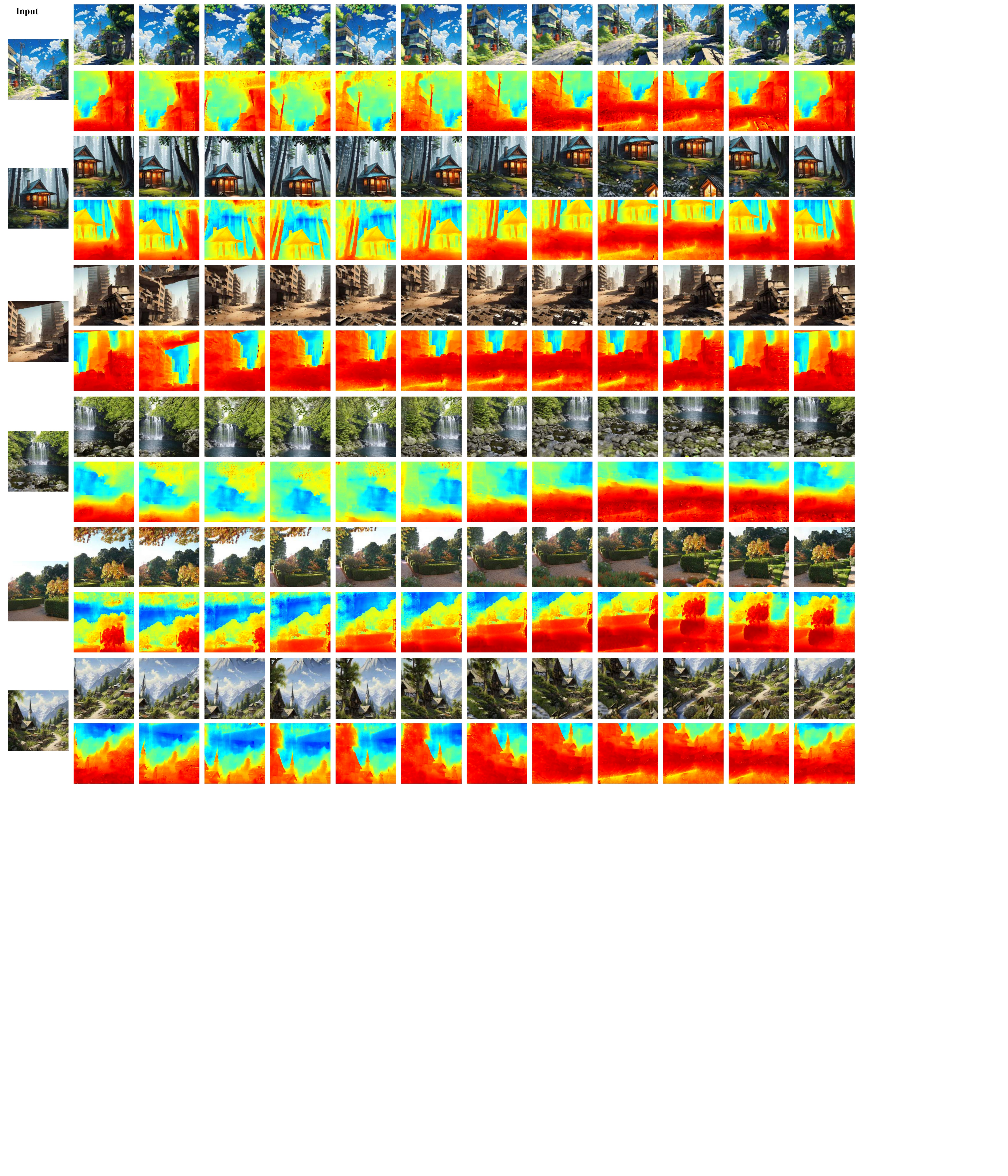}
   Part 2\,/\,4
  \caption{More qualitative examples generated by our Scene123 from a single image input.}
  \label{fig:supp_result2}
\end{figure*}

\begin{figure*}[hbtp]
  \centering
  \includegraphics[width=1\linewidth]{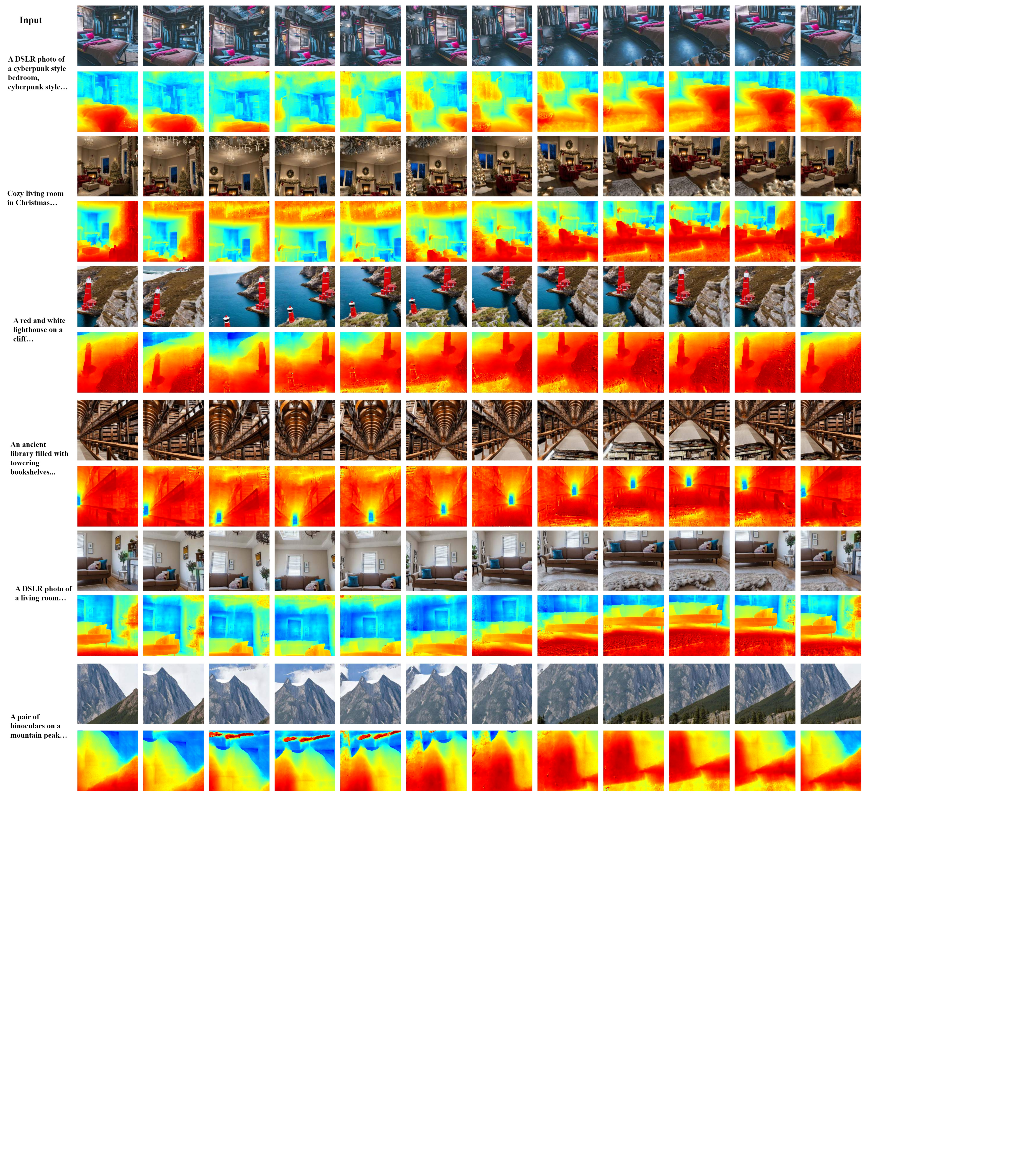}
   Part 3\,/\,4
  \caption{More qualitative examples generated by our Scene123 from a text prompt input.}
  \label{fig:supp_result3}
\end{figure*}

\begin{figure*}[hbtp]
  \centering
  \includegraphics[width=1\linewidth]{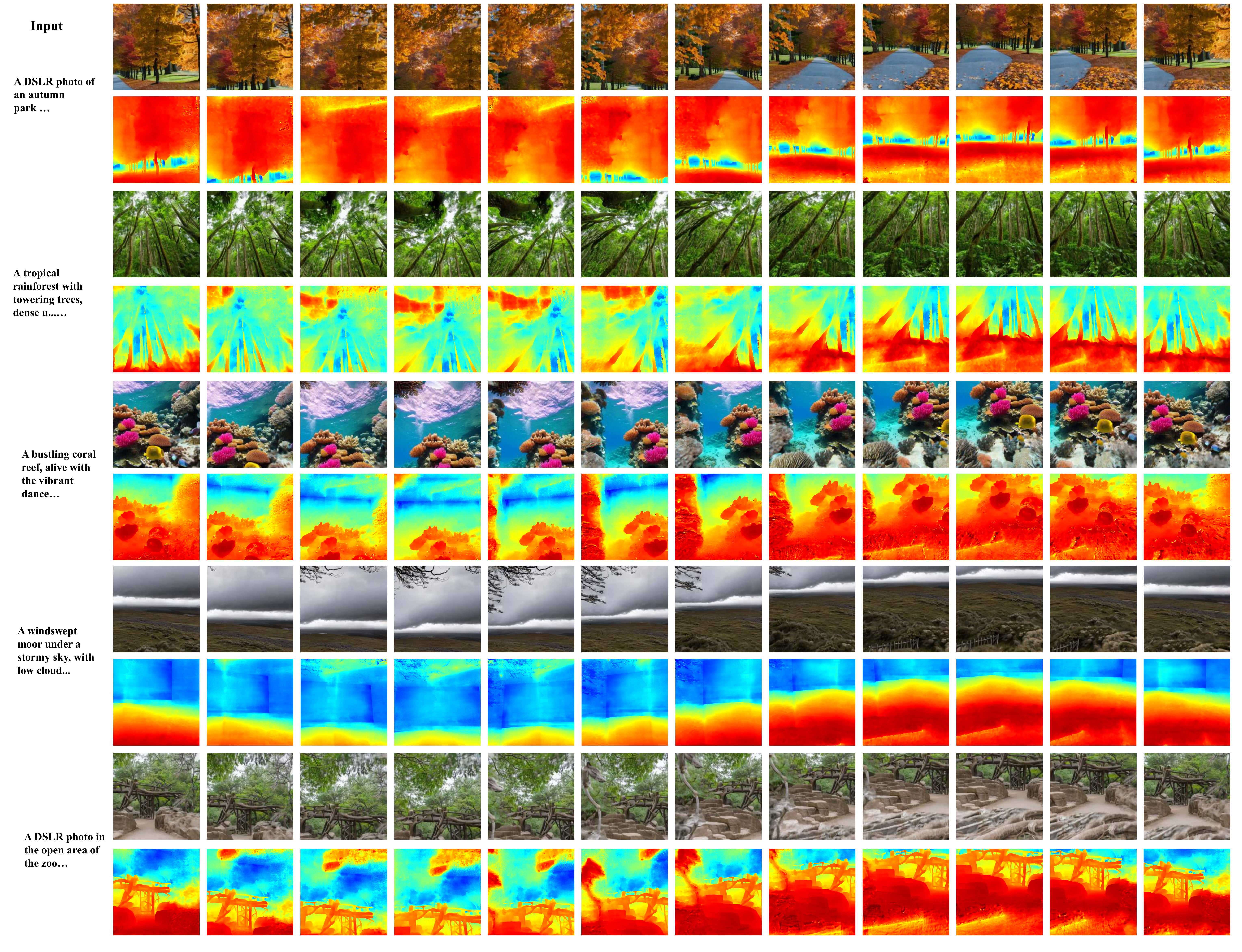}
   Part 4\,/\,4
  \caption{More qualitative examples generated by our Scene123 from a text prompt input.}
  \label{fig:supp_result4}
\end{figure*}

We provide more qualitative results in Fig.~\ref{fig:supp_result1}, Fig.~\ref{fig:supp_result2}, Fig.~\ref{fig:supp_result3}, Fig.~\ref{fig:supp_result4}, including indoor scenes, outdoor scenes, outdoor buildings, object-centered scenes with realistic renderings and precise depth details.

\end{document}